\documentclass[sigconf]{acmart}
\pdfoutput=1
\usepackage[ruled,vlined,linesnumbered]{algorithm2e}
\usepackage{bbding}
\AtBeginDocument{%
  \providecommand\BibTeX{{%
    \normalfont B\kern-0.5em{\scshape i\kern-0.25em b}\kern-0.8em\TeX}}}

\renewcommand\footnotetextcopyrightpermission[1]{}
\begin{document}
\fancyhead{}

\title{Dual-Level Decoupled Transformer for Video Captioning}

\author{Yiqi Gao$^{1,2}$ \quad Xinglin Hou$^{3}$ \quad Wei Suo$^{1,2}$ \quad Mengyang Sun$^{1,2}$ \\
\quad Tiezheng Ge$^3$ \quad Yuning Jiang$^3$ \quad Peng Wang$^{1,2}$} 

\affiliation{
 \institution{\textsuperscript{\rm 1}School of Computer Science, Northwestern Polytechnical University} 
 \institution{\textsuperscript{\rm 2}National Engineering Lab for Integrated Aero-Space-Ground-Ocean Big Data Application Technology}
 \institution{\textsuperscript{\rm 3}Alibaba Group}
 \country{}
 }
 
\email{{gyqjz, suowei1994, sunmenmian}@mail.nwpu.edu.cn}
\email{{xinglin.hxl, tiezheng.gtz, mengzhu.jyn}@alibaba-inc.com}
\email{peng.wang@nwpu.edu.cn}

\def\authors{Yiqi Gao, Xinglin Hou, Wei Suo, Mengyang Sun, Tiezheng Ge, Yuning Jiang and Peng Wang}
\renewcommand{\shortauthors}{Yiqi Gao, et al.}

\begin{abstract}
Video captioning aims to understand the spatio-temporal semantic concept of the video and generate descriptive sentences. The de-facto approach to this task dictates a text generator to learn from \textit{offline-extracted} motion or appearance features from \textit{pre-trained} vision models. However, these methods may suffer from the so-called \textbf{\textit{"couple"}} drawbacks on both \textit{video spatio-temporal representation} and \textit{sentence generation}. For the former, \textbf{\textit{"couple"}} means learning spatio-temporal representation in a single model(3DCNN), resulting the problems named \emph{disconnection in task/pre-train domain} and \emph{hard for end-to-end training}. As for the latter, \textbf{\textit{"couple"}} means treating the generation of visual semantic and syntax-related words equally. To this end, we present $\mathcal{D}^{2}$ - a dual-level decoupled transformer pipeline to solve the above drawbacks: \emph{(i)} for video spatio-temporal representation, we decouple the process of it into "first-spatial-then-temporal" paradigm, releasing the potential of using dedicated model(\textit{e.g.} image-text pre-training) to connect the pre-training and downstream tasks, and makes the entire model end-to-end trainable.
\emph{(ii)} for sentence generation, we propose \emph{Syntax-Aware Decoder} to dynamically measure the contribution of visual semantic and syntax-related words. Extensive experiments on three widely-used benchmarks (MSVD, MSR-VTT and VATEX) have shown great potential of the proposed $\mathcal{D}^{2}$ and surpassed the previous methods by a large margin in the task of video captioning.
\end{abstract}

\maketitle

\begin{figure}[ht!]
\includegraphics[width =0.48\textwidth]{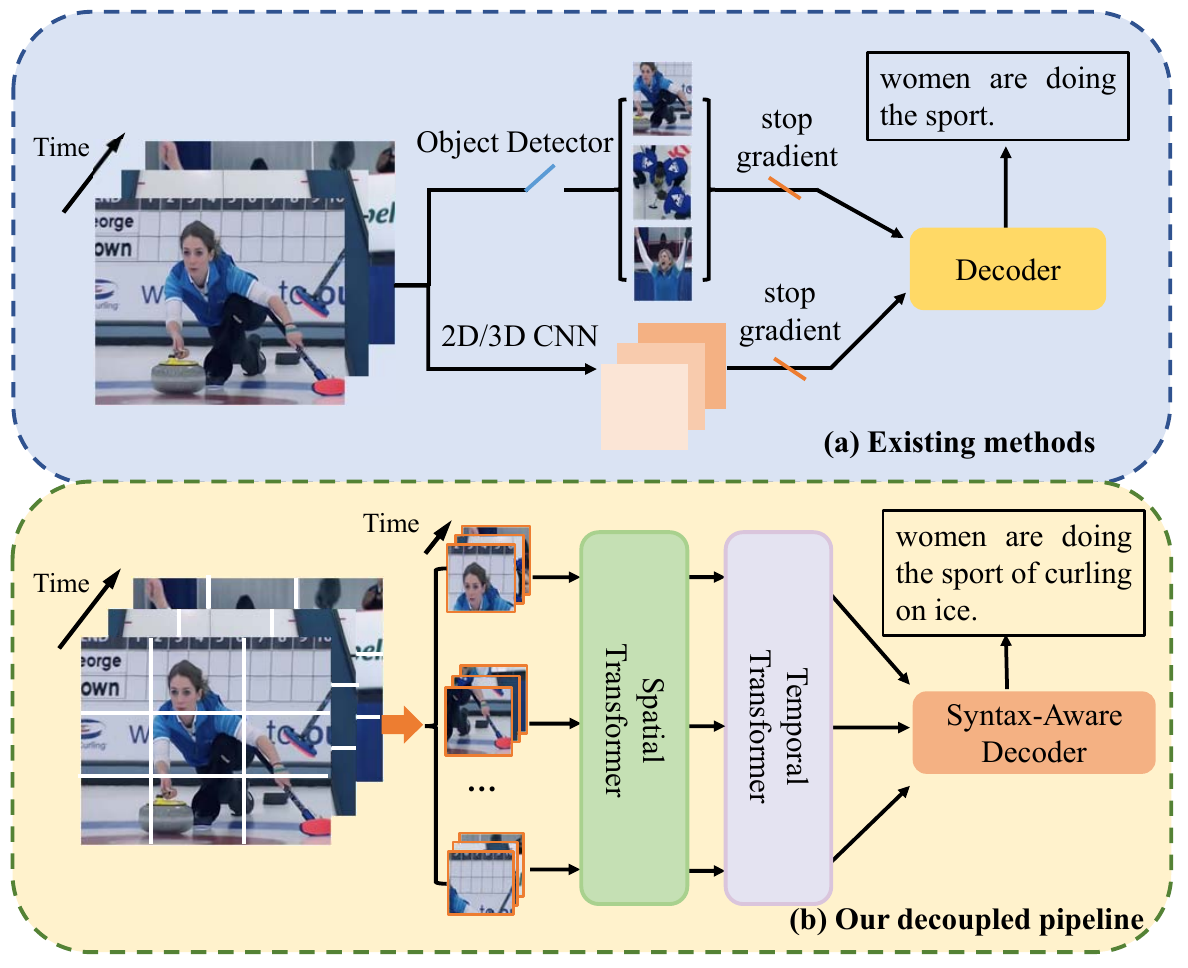}
\caption{\textit{Top}: The 3D-CNN or the 2D variant with complex temporal fusion block , along with an detection module to focus on the interested objects, in conventional two-stage video captioning model is generally infeasible for end-to-end finetuning due to the memory constraint on long video. \textit{Bottom}: The proposed $\mathcal{D}^{2}$ model overcomes this limit with the unified attention modules to align the language information and the decoupled spatial and temporal features which are represented by the residual coding. Benefiting from the modularized design, we can instantiate each block with dedicated pre-trained model and jointly optimize the entire system. Please refer to Fig~\ref{fig:model} for the detailed design.}
\label{fig1}
\end{figure}

\section{Introduction}

\textit{Video captioning}, which aims to understand spatio-temporal relation inside the video and describe it with natural language sentences, is a fundamental research task for multi-modal video-and-language understanding. It becomes an emerging requirement with the rapid emergence of videos in our lives. To generate good captions for videos, it involves not only the understanding of spatio-temporal semantics in videos but also expressing these factors into a natural language. Existing works~\cite{zhang2020object, zhang2019object, zhang2021open, aafaq2019spatio} mainly adopt a two-stage framework for video captioning: firstly extracting visual representation of the video using offline feature extractor (3DCNN and object detector), and then decode natural sentences based on these fixed features (Fig~\ref{fig1} (top)).

Despite being reasonable, these methods still suffer from the so-called \textbf{\textit{"couple"}} drawbacks on both video spatio-temporal representation and sentence generation process. For video spatio-temporal representation, \textbf{\textit{"couple"}} means that the learning process of spatio-temporal semantic is restricted in a single model(Fig~\ref{fig1}), \emph{i.e.}, 3D CNN, which brings two main limitations: \emph{(i)} disconnection in task/pre-train domain: offline feature extractors are often trained on tasks and domains different from the target tasks. For example, 3D CNN is generally trained from pure video data without any textual input on action recognition task, while being applied to video captioning. Large-scale video-text pre-training offers a way to mitigate this issue. However, compared with its image counterpart, the collection of the video-text dataset is much more complex and its noise is also much larger~\cite{miech2020end}. This makes the video-text pre-trained models are difficult to play a big role in the video captioning tasks. Besides, convolutional kernels are speciﬁcally designed to capture short-range spatio-temporal information, they cannot model long-range dependencies that extend beyond the receptive ﬁeld. \emph{(ii)}: end-to-end training: due to the memory and computation limitation, it is also infeasible to directly plug these feature extractors into a video captioning framework for end-to-end fine-tuning, causing the disconnection between pre-trained feature and downstream tasks; For sentences generation, \textbf{\textit{"couple"}} means that existing decoding methods pay equal attention to visual semantic(\emph{"woman"} in Fig~\ref{fig1}) and syntax-related words(\emph{"are"} in Fig~\ref{fig1}) during the whole sentence generation process, making the generation process unreasonable.

To tackle the above drawbacks, we propose $\mathcal{D}^{2}$, a dual-level decoupled pure transformer pipeline for end-to-end video captioning. In terms of video spatio-temporal representation, we \emph{decouple} the learning process into "first-spatial-then-temporal" paradigm. technically, we firstly use a 2D vision transformer to generate a spatial representation for each frame, then, we propose a \emph{Residual-Aware Temporal Block(RATB)} to build the temporal relationship between each frame. This brings us two main advantages over the two limitations mentioned above: \emph{(i)}: compared to the video-text dataset, image-text dataset is easy to access, releasing the potential of applying image-text pre-training to our 2D vision transformer, which is more suitable to multimodal tasks like video captioning. \emph{(ii)} due to the lightweight nature of the 2D model (as opposed to 3D CNN), we can easily perform end-to-end training, building the connection between pre-trained feature and downstream video captioning tasks. In terms of sentence generation, we \emph{decouple} the generation process of visual semantic and syntax-related words. Concretely, using the syntactic prior provided by the pre-trained language model, our \emph{Syntax-Aware Decoder(SAD)} dynamically measures the contribution of visual features and syntactic prior information for the generation of each word, facilitating a more reasonable and fine-grained video captioning generation. 

Experimentally, we conduct a series of ablation studies on different modules for \emph{decouple} spatio-temporal representation learning, as well as modules for \emph{decoupling} visual semantic and syntax-related words generation, gaining insights on the performance of our novel \emph{decouple} pipeline in video captioning task. Our \emph{$D^{2}$}, when tested on the three benchmarks(MSRVTT, MSVD and VATEX), outperforms existing methods on all metrics by a large margin. 
Our main contributions are summarized as follows.
\begin{itemize}
\vspace{-1.5mm}
\item 
We propose $\mathcal{D}^{2}$, a novel transformer pipeline, which decouple the process of \emph{video spatio-temporal representation learning} and \emph{sentence generation}. For video spatio-temporal representation learning, our pipeline decouple the previous offline spatio-temporal representation learning  into \textbf{\textit{"first spatial then temporal"}} paradigm, addressing the problems of \emph{disconnection in task/pre-train domain} and \emph{end-to-end training}.
\item 
For caption generation, our model decouples the generation of visual semantic and syntax-related words, adaptively measuring its contribution, resulting in a more reasonable and fine-grained video captioning generation process.
\item 
We show that $\mathcal{D}^{2}$ surpasses all previous methods for video captioning, achieving a new state-of-the-art on MSVD, MSR-VTT and VATEX benchmarks.
\end{itemize}

\section{Related Work}
\subsection{Image and Video Representation Learning With Transformer}
Inspired by the development of natural language processing, the proposed ViT~\cite{dosovitskiy2020image} firstly introduces transformer into image classification and achieves surprising performance. This motivates many researchers to conduct a more in-depth study of the transformer as the backbone network. Different from CNNs, transformers are not limited by the receptive field and can obtain more comprehensive contextual information. Meanwhile, due to the characteristics of the attention mechanism to dynamically generate attention coefficients for different instances, the expressive ability of transformers is also stronger. Considering the above advantages, researchers are increasingly applying transformers to extract feature representations, both in the field of images and videos. For the image domain, multi-scale features~\cite{Wang_2021_ICCV,Yuan_2021_ICCV, liu2021swin} are expanded by introducing pyramid structure into the transformer, allowing the transformer-based backbone to better adapt to the downstream vision tasks. Several work~\cite{liu2021swin,yuan2021volo,chu2021Twins} also focuses on the balance between attention span and computational overhead by adding local windows or filtering high-value patches.

When considering the video domain, timing information should be added. TimeSformer~\cite{bertasius2021space} studies five different variants of space-time attention and suggests a factorized space-time attention for its strong speed-accuracy trade-off. ViViT~\cite{arnab2021vivit} examines four factorized designs of spatial and temporal attention for the pre-trained ViT model. Similar to our work, they adopt divided attentions on spatial and temporal with two-path transformer models. However, they focus on patch-level attention for video action recognition. We mainly investigate frame-level spatio-temporal semantic representation for video captioning tasks. Video Swin-Transformer~\cite{liu2021video} designs a video recognition framework by transferring the thought of swin-transformer from space domain to space-time domain. Although this method has achieved certain results, the design of 3D local window still makes the overall framework complex.

The training of backbone networks often needs to be driven by a large amount of data, which usually consumes a lot of computing resources. In this paper, we design an end-to-end framework to avoid the problem of offline characterization, making the pipeline more concise.

\subsection{Video Captioning}
Over the past years, we have witnessed the great development of video captioning. Before the emergence of neural networks, template-based methods that used the concepts of Verb, Subject and Object (SVO)~\cite{barbu2012video,kojima2002natural} have become dominant. While in the era of deep learning, a broad collection of methods have been proposed ~\cite{venugopalan2014translating, ryu2021semantic} which mainly adopt an encoder-decoder framework. These works tend to extract images' features with CNNs and adopt RNNs to generate descriptions. Venugopalan~\cite{venugopalan2014translating} firstly introduces an LSTM to generate the mean pooled representations over all frames. Wang~\cite{wang2018reconstruction} tries to enhance the quality of generated captions by reproducing the frame features from decoding hidden states. And Buch~\cite{buch2017sst} connects the same output from the two hidden layers of the opposite direction at a particular time step to further refine the descriptions. However, the original implementation equipped with RNN is difficult to capture long-term dependencies since only the last hidden state takes part in the final generation. The proposed attention mechanism~\cite{gao2017video, hori2017attention, tu2017video, yao2015describing} alleviates this deficiency. For each output that the decoder generates, it has access to the entire input sequence at the temporal level. Note that different regions of each video frame also have different contributions to the prediction of the final word. For example, the description object is more important than the background. Therefore, it is also necessary to introduce spatial attention. Li~\cite{li2017mam} adopts two layers of spatio-temporal dynamic attention for video subtitles. When calculating the spatial attention weight of a specific frame, it will also consider the attention weight of the previous frame. In this way, the spatial attention map is linked across time. Wang~\cite{wang2018spotting} tries to learn a model to distinguish the foreground from the background in the video without explicit supervision, and calculates the significance score from the spatial feature map to separate the foreground and background according to the threshold. Recent work typically follows the pipeline where off-the-shelf 2D and 3D is used to extract spatio-temporal feature for video representation. Unlike previous works, we mainly focus on decouple the spatio-temporal representation learning process for better learning.
\section{Our Approach}

\begin{figure*}[t]
\includegraphics[width =1\textwidth]{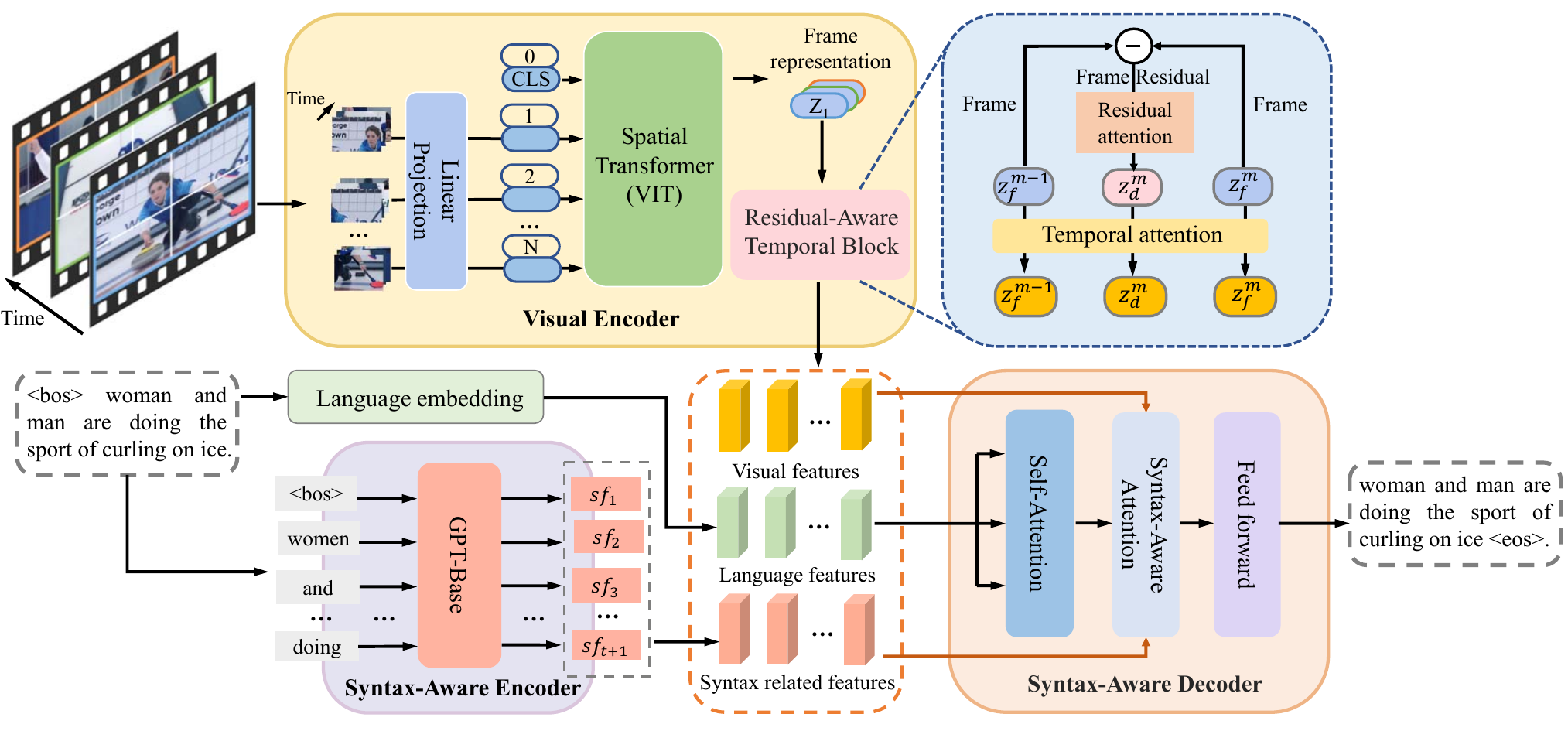}
\caption{The overall architecture of our proposed $\mathcal{D}^{2}$. $\mathcal{D}^{2}$ decouples the learning process in both video and caption side. On the video side, spatial representation is obtained using \textit{Spatial Transformer(ViT)}. With a carefully designed \textit{Residual-Aware Temporal Block}, temporal dependency is built both on \textit{residual} and \textit{long-range} level. On the caption generation side, $\mathcal{D}^{2}$ decouple the generation of visual semantic and syntax-related words by a \textit{Syntax-Aware Decoder}, in which semantic and syntactic feature is fused dynamically during the whole generation process. Benefiting from the \textit{neat} and \textit{modularized} design, we can instantiate each module with dedicated pre-trained model and jointly optimize the entire system for task-specific fine-tuning.}
\label{fig:model}
\end{figure*}

In this section, we introduce the proposed video captioning framework $\mathcal{D}^{2}$, which is illustrated in Figure~\ref{fig:model}. Compared with existing method, $\mathcal{D}^{2}$ decouple the video captioning process both in video representation and caption generation phase.  We begin with introducing the \textit{spatial encoder}. And then, the \textit{Residual-Aware Temporal Block}(RATB) and the \textit{Syntax-Aware Decoder}(SAD) are elaborated. The training details will be given at the end of this section. 

\subsection{Spatial Encoder}~\label{3.1}
Given a video $\mathcal{V}$, the video captioning task aims to generate a caption $y=\left\{y_{1}, \cdots, y_{T}\right\}$ to describe semantic concepts in $\mathcal{V}$, where $y_{t}$ denotes the $t$-th word in the caption. In the following, we introduce how a frame representation is obtained through a spatial encoder. 

The video clips $v_{i}\in\mathcal{V}$ are represented as a sequence of frames (images) in our paper. Specifically, the video clip $v_{i}$ is composed of $\left|v_{i}\right|$ sampled frames such that $\left|v_{i}\right|=\left\{v_{i}^{1}, v_{i}^{2}, \ldots, v_{i}^{\left|v_{i}\right|}\right\}$. Unlike previous methods~\cite{zhang2021open, zhang2019object, wang2019controllable, zhang2020object} which extract clip features using pre-trained CNN~\cite{carreira2017quo,he2016deep} or object detector~\cite{ren2016faster}. Our $\mathcal{D}^{2}$ model is trained on pixels directly via taking the frames as input in an end-to-end manner. 

In order to get the video representation, we first extract the frames from each video clip and then encode them through our spatial encoder to get a sequence of frame features. In this paper, we adopt a 2D ViT-B/16~\cite{dosovitskiy2020image} as our spatial encoder and mainly consider using a sequence of frame representation as video representation. The ViT first extracts non-overlapping image patches, then performs a linear projection to project them into 1D tokens. The transformer architecture is used to model the interaction between each patch of the input frame to get the final representation. We use the output from the [CLS] token as the frame representation. For the input frame sequence of video $v_{i}=\left\{v_{i}^{1}, v_{i}^{2}, \ldots, v_{i}^{\left|v_{i}\right|}\right\}$, the generated representation can denote as $\mathbf{Z}_{i}=\left\{\mathbf{z}_{i}^{1}, \mathbf{z}_{i}^{2}, \ldots, \mathbf{z}_{i}^{\left|v_{i}\right|}\right\}$. Benefiting from the modularized design of model, we can instantiate our spatial encoder with dedicated image-text pre-trained model. specifically, we utilize the recent one-stage image-text pre-trained model ALBEF~\cite{li2021align} to initialize our spatial encoder. The impact of different weight initialization strategies is examined in our ablation.

\subsection{Residual-Aware Temporal Block} 
\label{section:difference}
Considering that viewing video representation as a sequence of frame representations may ignore temporal dependency, we propose Residual-Aware Temporal Block(RATB) to build both long-range and Residual level temporal dependency between each frame. Since two successive frames contain content displacement, which reﬂects the actual actions, we explicitly propose a Residual Attention Mechanism to extend the input frame representation and guide the temporal transformer to encode more motion-related representations. Speciﬁcally, we adopt transformed residual of frame representation between adjacent time stamps to describe the motion change, which is formulated as:

\begin{equation}
    \mathbf{Z}_{\mathbf{d}}=\delta\left(\left\{Z_{f}^{1}-Z_{f}^{0}, Z_{f}^{2}-Z_{f}^{1}, \ldots, Z_{f}^{m-1}-Z_{f}^{m-2}\right\}+\mathbf{P}\right)
\end{equation}
where $\mathbf{P}$ is the positional embedding, $Z_{f}^{m-1}$ and $Z_{f}^{m}$ are two adjacent frame representations, $\delta$ is 1-layer transformer encoder layer, and $\mathbf{Z}_{\mathbf{d}}$ is the difference representations. We insert difference representations between every frames as below:
\begin{equation}
    \mathbf{Z}_{\mathbf{n}}=\left\{Z_{f}^{0}, Z_{d}^{1}, Z_{f}^{1}, Z_{d}^{2}, Z_{f}^{2}, \ldots, Z_{d}^{m-1}, Z_{f}^{m-1}\right\}+\mathbf{P}+\mathbf{T}
\end{equation}
where $\mathbf{Z}_{\mathbf{n}}$ is the output of our RATB, $\mathbf{P}$ is the positional embedding, $\mathbf{T}$ is the type embedding. 
With the design of residual-level attention and frame-level temporal attention, atomic actions in short term segments can be contextualized with the rest of the video, thus to be fully disambiguated. 

Compared with the 3D-CNN counterpart, which is inherently limited in capturing long-range dependencies by means of aggregation of shorter-range information, our RATB built temporal dependency at both shorter-range and long-range levels.
\subsection{Syntax-Aware Decoder}
Existing decoding methods in video captioning were suffering from the semantics and syntax coupling drawback. $\mathcal{D}^{2}$  solve this problem by dynamic fusing semantic features and syntactic features during decoding. We begin with the syntactic feature extraction and then introduce the semantic and syntactic decoding method.
\subsubsection{Syntactic Feature Extraction}
Unlike previous works using part-of-speech(POS)~\cite{wang2019controllable} tagging tools, we choose a pre-trained language model to provide syntactic prior information. To be specific, in order to get syntactic features during the decoding stage, we build a pretrained 12-layers GPT-based language model to extract syntactic features. The language model was first pre-trained by using a large corpus of documents, and then fine-tuned by using captions sentences only. Some previous works~\cite{clark2019does} have proven that the pre-trained language model could retain syntactic information inside some of their attention heads. So in that way, we could obtain dense syntactic features containing not only POS tagging information but also the whole syntactic information by using the pretrained language model.

Specifically, given the word sequence $Y$ = $\left(<\right.$bos$\left.>, {y}_{1}, {y}_{2}, \ldots, {y}_{N} \right)$, our syntax-aware encoder is asked to predict this offset sequence $\hat{Y}=\left(\hat{y}_{1}, \hat{y}_{2}, \ldots, \hat{y}_{N},<\right.$eos$\left.>\right)$ by one forward process. This entire process can be expressed as follows:
\begin{equation}
    sf=\text {GPT}(Y+pos)
\end{equation}
\begin{equation}
    \hat{Y}=\log _{-} \operatorname{softmax}(FF(sf))
\end{equation}
where $pos$ is position embedding, $FF$ is feedforward network, $\hat{Y}$ is the output distribution of predicted words, $sf$ is our syntactic feature. This language model is trained with XE loss. This can be expressed as:
\begin{equation}
    sf_{t} \leftarrow SAE\left(Y_{<t}\right), sf_{t} \in \mathbb{R}^{d_{\text {model}}}
\end{equation}
where $\mathbf{SAE}$ is our Syntax-Aware Encoder.
\subsubsection{Visual Semantic and Syntactic Decoding}
After extracting syntactic features, we combine it with the output of our RATB as our candidates during decoding. In our opinion the visual encoder could provide more semantic information, so the decoder could choose semantic or syntactic clues flexibility by using the attention mechanism. Besides, we combine the visual encoder output features, the pretrained language model output features and the current word embedding together as the query vector of the attention module. By this way, we hope the decoder could avoid generating trivial words. This can be formulated as follows:
\begin{equation}
h_{t}=\operatorname{Decoder}\left(\left[\mathbf{Z}_{\mathbf{n}} ; sf_{t}\right], Y_{<t}\right)
\end{equation}
\begin{equation}
    q_{i, t}=h_{t} W_{i}^{Q}, k_{i,t}= \left[\mathbf{Z}_{\mathbf{n}} ; sf_{t}\right] W_{i}^{K}, v_{i,t}= \left[\mathbf{Z}_{\mathbf{n}} ; sf_{t}\right]  W_{i}^{V}
\end{equation}
\begin{equation}
    \text {head}_{i, t}=\operatorname{softmax}\left(q_{i, t} k_{i, t}^{T}\right) v_{i, t}
\end{equation}
\begin{equation}
    \text {head}_{i}=\text {Concate}\left(\text {head}_{i,1}, \ldots, \text {head}_{i,M}\right)
\end{equation}
\begin{equation}
    \text {att}=\text {Concate}\left(\text {head}_{1}, \ldots, \text {head}_{h}\right) W^{O}
\end{equation}
where $\mathbf{Z}_{\mathbf{n}}$ is the output of our RATB, $sf_{t}$ is the output of syntactic encoder, $q_{i, t}$, $k_{i,t}$, $v_{i,t}$ are the query, key matrix and value matrix for the t time step word in head i respectively, ${head}_{i, t}$ is the attention result at t-th timestep, ${head}_{i}$ is the attention result in head i, att is the final attention coefficient for sequence generation.

\subsection{Training details}
The captioning model is typically trained by the cross-entropy loss (XE) given the ground-truth pair $\left(\mathcal{V}, y^{*}\right)$.

\begin{equation}
       L_{X E}(\theta)=-\sum_{t=1}^{T} \log \left(p_{\theta}\left(y_{t}^{*} \mid y_{1: t-1}^{*}, \mathcal{V}\right)\right),
\end{equation}

where $\theta$ is the parameters of our model, $y_{1: T}^{*}$ is the target ground truth sequence.

To address the exposure bias and target mismatch problem in XE, we directly optimize the non-differentiable metric with Self-Critical Sequence Training~\cite{rennie2017self}:

\begin{equation}
    L_{R L}(\theta)=-E_{y_{1: T} p_{\theta}}\left[r\left(y_{1: T}\right)\right],
\end{equation}

where the reward $r(\cdot)$ is the CIDEr-D score.

Besides, following~\cite{cornia2020meshed}, we use the mean of rewards rather than greedy decoding to baseline the reward. The gradient expression for one sample is formulated as:
\begin{equation}
    b=\frac{1}{k}\left(\sum_{i}^{k} r\left(y_{i}\right)\right),
\end{equation}
\begin{equation}
    \nabla_{\theta} L_{R L}(\theta) \approx-\frac{1}{k} \sum_{i=1}^{k}\left(\left(r\left(y_{1: T}^{i}\right)-b\right) \nabla_{\theta} \log p_{\theta}\left(y_{1: T}^{i}\right)\right),
\end{equation}
where k is the number of the sampled sequences, $y_{1: T}^{i}$ is the $i\text{-th}$, and b is the mean of the rewards obtained by the sampled sequences.

\section{Experiments}
In this section, we conduct experiments to verify the effectiveness of the proposed methods. Firstly, We introduce three widely-used datasets: MSVD~\cite{chen2011collecting}, MSRVTT~\cite{xu2016msr} and recent VATEX~\cite{wang2019vatex}. Then implementation details including hyper-parameters and techniques are illustrated. After that, we make comparisons between our methods and the state-of-the-arts. More ablation studies are also discussed in the final part.

\begin{table*}[htbp]
\begin{tabular}{ l c cc cccc cccc}
\toprule[1pt]
Model  & Ref & 3DCNN & Detector& \multicolumn{4}{c}{MSVD} & \multicolumn{4}{c}{MSRVTT} \\
~ & ~ & ~ &~ & B@4 & M &R &C &B@4 & M &R &C \\
\hline
MGSA~\cite{chen2019motion} & AAAI19 & \checkmark &\small\XSolid &  53.4   & 35.0   & -       & 86.7     & 45.4   & 28.6   & -     & 50.1    \\
FCVC~\cite{fang2019fully}  & AAAI19 & \checkmark &\small\XSolid & 53.1   & 34.8   & 71.8    & 79.8     & -      & -      & -     & -       \\
POS-CG~\cite{wang2019controllable}  & ICCV19 & \checkmark &\small\XSolid & 52.5   & 34.1   & 71.3    & 88.7     & 42.0   & 28.2   & 61.6  & 48.7    \\
POS-CG+RL~\cite{wang2019controllable}       & ICCV19      & \checkmark& \small\XSolid& 53.9   & 34.9   & 72.1    & 91.0     & 41.3   & 28.7   & 62.1  & 53.4    \\
POS-VCT~\cite{hou2019joint}  & ICCV19 & \checkmark &\small\XSolid & 52.8   & 36.1   & 71.8    & 87.8     & 42.3   & 29.7   & 62.8  & 49.1    \\
MARN~\cite{pei2019memory} & CVPR19 & \checkmark &\small\XSolid & 48.6   & 35.1   & 71.9    & 92.2     & 40.4   & 28.1   & 60.7  & 47.1    \\
PMI-CAP~\cite{chen2020learning}  & ECCV20 & \checkmark &\small\XSolid & 54.7   & 36.4   & -       & 95.2     & 42.1   & 28.7   & -     & 49.4    \\
PMI-CAP+Audio~\cite{chen2020learning}    &  ECCV20    & \checkmark&\small\XSolid & -      & -      & -       & -        & 43.9   & 29.5   & -     & 50.6    \\
SGN~\cite{ryu2021semantic}       & AAAI21 & \checkmark &\small\XSolid & 52.8 & 35.5 & 72.9    & 94.3     & 40.8   & 28.3   & 60.8  & 49.5    \\
NACF~\cite{yang2021non}          & AAAI21 & \checkmark &\small\XSolid & 55.6 & 36.2 & - & 96.3 & 42.0 & 28.7 & - & 51.4    \\ \hline

OA-BTG~\cite{zhang2019object}    & CVPR19 & \checkmark &\checkmark & 56.9 & 36.2 & -     & 90.6  & 41.4 & 28.2 & -     & 46.9  \\
GRU-EVE~\cite{aafaq2019spatio}   & CVPR19 & \checkmark &\checkmark & 47.9 & 35.0 & 71.5  & 78.1  & 38.3 & 28.4 & 60.7  & 48.1  \\
STG-KD~\cite{pan2020spatio}      & CVPR20 & \checkmark &\checkmark & 52.2 & 36.9 & 73.9  & 93.0  & 40.5 & 28.3 & 60.9  & 47.1  \\
SAAT~\cite{zheng2020syntax}      & CVPR20 & \checkmark &\checkmark & 46.5 & 33.5 & 69.4  & 81.0  & 40.5 & 28.2 & 60.9  & 49.1  \\
ORG-TRL~\cite{zhang2020object}   & CVPR20 & \checkmark &\checkmark & 54.3 & 36.4 & 73.9  & 95.2  & 43.6 & 28.8 & 62.1  & 50.9  \\
OPEN-BOOK~\cite{zhang2021open}   & CVPR21 & \checkmark &\checkmark & -    & -    & -     & -     & 42.8 & 29.3 & 61.7  & 52.9  \\
O2NA~\cite{liuo2na}              & ACL21  & \checkmark &\checkmark & 55.4 & 37.4 & 74.5  & 96.4  & 41.6 & 28.5 & 62.4  & 51.1  \\ 
\hline
$\mathcal{D}^{2}$  & -      & \small\XSolid &\small\XSolid & {\bf56.9}  &{\bf38.4}     &{\bf75.1}  &{\bf99.2}   & {\bf44.5} & {\bf30.0}    & {\bf63.3}  & {\bf56.3}     \\ 
\toprule[1pt]
\end{tabular}
\caption{\textbf{Performance comparisons on MSVD and MSRVTT benchmarks.} B@4,C,M,and R denote BLEU@4, CIDEr-D, METEOR, and ROUGE\_L, respectively. - means not available.}
\label{msrvtt}
\end{table*}
\subsection{Datasets}
\subsubsection{MSVD}
The MSVD dataset contains 1970 video clips and roughly 80000 English sentences. It was firstly developed in 2010 and used to test and train the translational and paraphrase algorithms. Similar to the prior work~\cite{wang2019controllable}, we separate the dataset into 1,200 train, 100 validation ,and 670 test videos.

\subsubsection{MSR-VTT}
The MSR-VTT dataset contains 10,000 video clips from the YouTube website. we follow the standard splits in~\cite{xu2016msr} for fair comparison which separates the dataset into 6,513 training, 497 validation and 2,990 test videos.

\subsubsection{VATEX}
The VATEX dataset is the most recently released large-scale dataset that contains 41,269 videos. Each video is annotated with 10 English and 10 Chinese descriptions. We utilize English corpora in our experiments. According to the ofﬁcial splits~\cite{wang2019vatex}, the dataset is divided into 25,991 training, 3,000 validation, and 6,000 public testing.

\subsection{Implementation Details.}
For the sentences longer than 30 words are truncated (50 for VATEX); the punctuations are removed (for VATEX are retained); all words are converted into lower case. 

Our model is trained on pixels directly via taking the frames as input. We divide video into 12 clips. If not otherwise stated, we randomly sample a single frame from each clip for training, and use the middle frame for inference. The impact of different number of clips is examined in ablation. The whole framework is optimized with ADAM~\cite{kingma2014adam} optimizer. We set the initial learning rate of the RATB and SAD to $10^{-4}$, the spatial encoder to $10^{-5}$. We pre-train our 2D visual encoder with the similar spirit as~\cite{li2021align}. For the other components, the parameters are randomly initialized with Xavier init.
Our training is started with cross entropy optimization. If the cider value drops for 5 consecutive epochs, it will turn to self-critical sequence training. The training process stops when the cider value drops for 5 consecutive epochs in self-critical sequence training.

We use input image size $256\times256$. Our model is implemented in PyTorch~\cite{paszke2019pytorch} and transformers~\cite{wolf2020transformers}.
We set the batch size to 48. The dimension of our spatial encoder is 768. We set the dimension of Residual-Aware Temporal Block and Syntax-Aware Decoder to 512. To avoid over-ﬁtting, we exploit dropout operation after the multi-head self-attention layer and the FFN of each transformer encoder layer. The dropout ratio is set to 0.1 by default. 

\begin{figure}[t]
\includegraphics[width =0.48\textwidth]{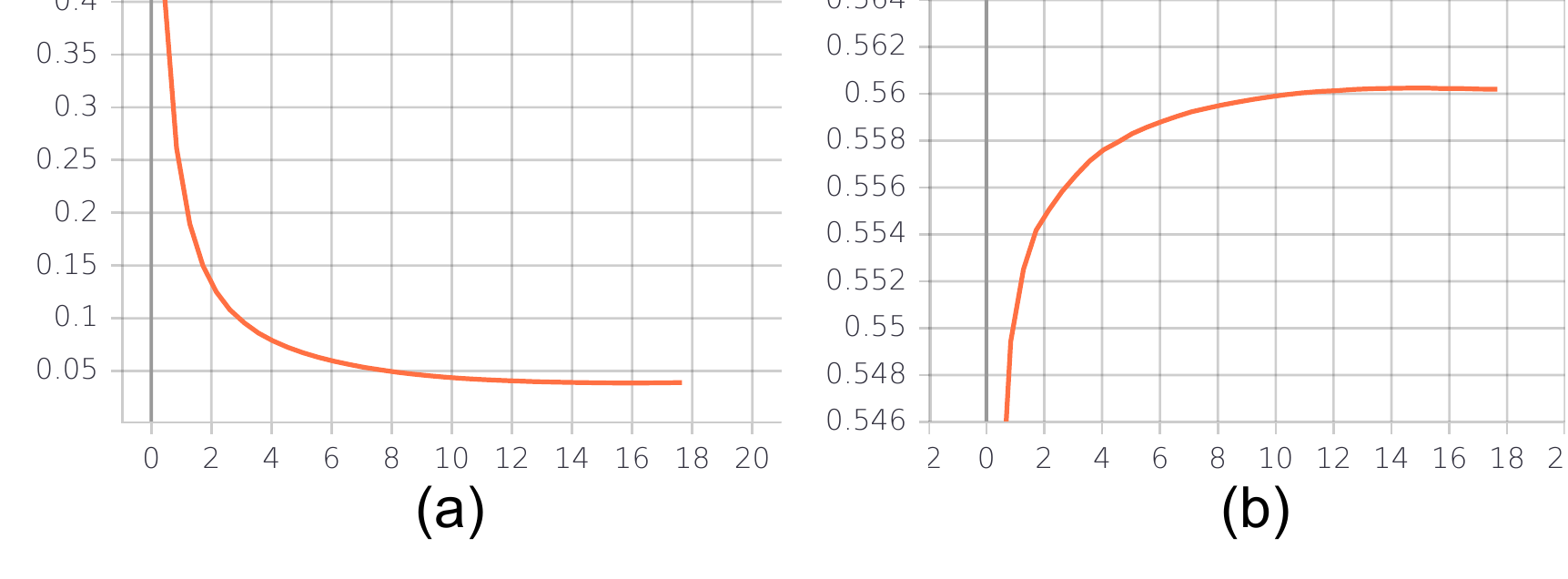}
\caption{\textit{(a)}: Validation loss during Syntax-Aware Encoder fine-tuning. \textit{(b)}: Accuracy on validation set.}
\label{fig:loss}
\end{figure}

To guarantee the quality of generated \textit{"syntax realted features"}, we build our \textit{Syntax-Aware Encoder} based on GPT-Base provided by pytorch-transformers~\cite{wolf2020transformers}. We then fine-tune our \textit{Syntax-Aware Encoder} on caption sentences to obtain syntactic information. Fig~\ref{fig:loss} shows the validation loss and validation accuracy on VATEX datasets, demonstrating the ability of our \textit{Syntax-Aware Encoder} to encode syntactic information.
\subsection{Comparison to State-of-the-Arts}
We compare our model with the following state-of-the-art method, which all use offline feature-extractor (either 3DCNN or object detector) for spatio-temporal representation.

\begin{itemize}
    \item 
    MGSA~\cite{chen2019motion}: LSTM-based model for motion-guided caption generation.
    \item
    FCVC~\cite{fang2019fully}: Fully convolutional network for coarse-to-fine caption generation.
    \item
    POS-CG~\cite{wang2019controllable}: Part-of-speech guided caption generation. 
    \item
    POS-VCT~\cite{hou2019joint}: This method takes pos for caption generation.
    \item
    MARN~\cite{pei2019memory}: This method leverages memory network to capture cross-video contents.
    \item
    PMI-CAP~\cite{chen2020learning}: This method learns pairwise modality interactions to better exploit complementary information for each pair of modalities in video.
    \item
    SGN~\cite{ryu2021semantic}: This method uses a semantic grouping network to capture the most discriminating word phrases.
    \item
    NACF~\cite{yang2021non}: This method uses a coarse-to-fine decoding method to better capture visual words in video.
    \item
    OA-BTG~\cite{zhang2019object}: This method uses a object-aware aggregation bidirectional temporal graph (OA-BTG) to capture detailed temporal dynamics for salient objects in video. 
    \item
    GRU-EVE~\cite{aafaq2019spatio}: A LSTM based model which capture spatial-temporal dynamics by Short Fourier Transform.
    \item
    STG-KD~\cite{pan2020spatio}: This method distills the knowledge of spatial-temporal object interactions from a spatial-temporal graph.
    \item
    SAAT~\cite{zheng2020syntax}: This method learns actions by simultaneously referring to the subject and video dynamics.
    \item
    ORG-TRL~\cite{zhang2020object}: This method uses a object relation graph to build the object interactions.
    \item    OPEN-BOOK\cite{zhang2021open}:They use ``retrieval and copy" pipeline to help caption generation. 
    \item   O2NA~\cite{liuo2na}: This method uses a Object-Oriented Non-\\Autoregressive approach (O2NA). 
    
\end{itemize}
\begin{table}[ht]
\centering
    \begin{tabular}{lccccc}
    \toprule[1pt]
    Model & B@4 & M & R & C \\ \hline
    VATEX (ICCV19)~\cite{wang2019vatex} & 28.7   & 21.9   & 47.2    & 45.6    \\
    ORG-TRL (CVPR20)~\cite{zhang2020object}  & 32.1   & 22.2   & 48.9    & 49.7    \\
    NSA (CVPR20)~\cite{guo2020normalized} & 31.0   & 22.7   & 49.0    & 57.1    \\
    OPENBOOK (CVPR21)~\cite{zhang2021open} & 33.9   & 23.7   & 50.2    & 57.5    \\ \hline
    $\mathcal{D}^{2}$ &35.1   &{\bf25.1}  &{\bf51.3}  &  {\bf60.9}            \\ \toprule[1pt]
    \end{tabular}
\caption{\textbf{\textit{Performance comparisions on VATEX benchmark}}. B@4, C, M, R denote BLEU@4, CIDEr-D, METEOR, and ROUGE\_L, respectively. }
\label{vatex}
\end{table}

\begin{figure*}[t]
\includegraphics[width =0.95\textwidth]{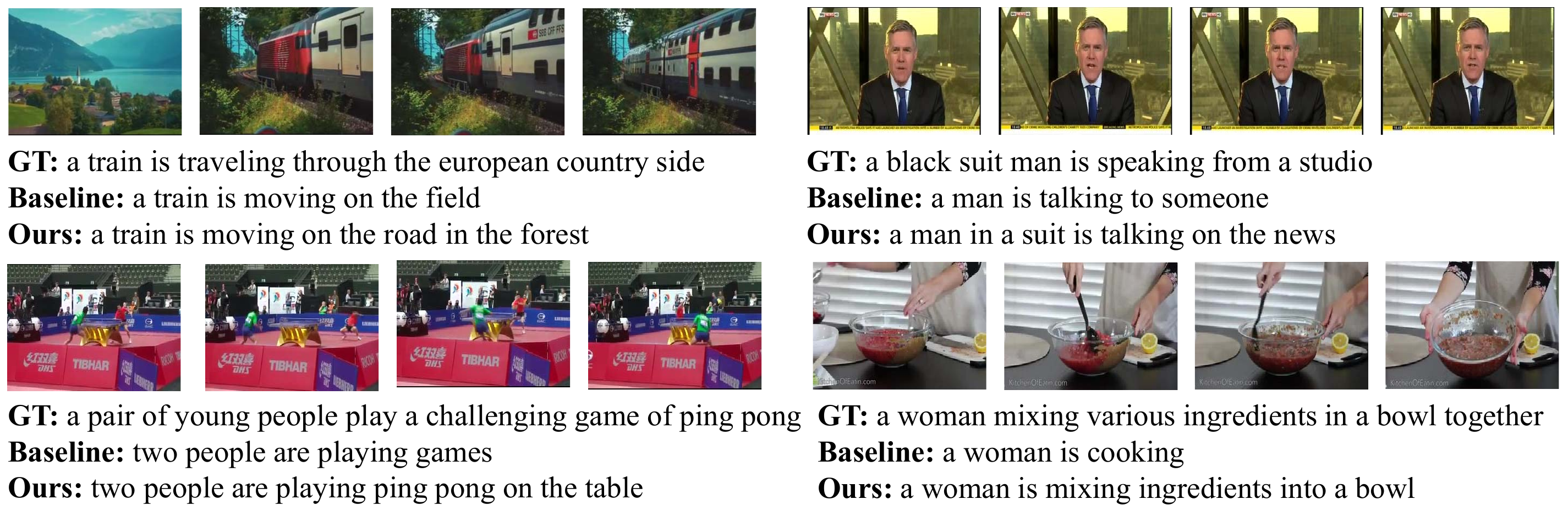}
\caption{\textbf{\textit{Visualization of generations on MSR-VTT with GT and our $\mathcal{D}^{2}$}}. Compraed with baseline, our $D^{2}$ can generate more richer and descriptive captions.}
\label{showcase}
\end{figure*}

Table~\ref{msrvtt} reports the video caption performances of different models on the MSVD and MSR-VTT datasets. We can see that our $D^{2}$ consistently exhibits better performance than the others. To be specific, on MSVD, our $D^{2}$ surpasses all the other methods for all metrics by a large margin even for the strongest competitor O2NA~\cite{liuo2na}. The CIDEr-D score of our method reaches 99.2\%, which advances O2NA~\cite{liuo2na} by 2.9\%. As for MSRVTT, our $D^{2}$ achieves better performance than all the other methods for all metrics. Compared with the strongest competitor OPENBOOK~\cite{zhang2021open}, we advance it by 6.4\%(56.3 \emph{vs.} 52.9). The boost of performance on both MSVD and MSRVTT demonstrates the advantages of our $D^{2}$ which uses \emph{decoupled} pipeline on both video spatio-temporal representation learning and sentence generation. Importantly, since CIDEr-D weights the n-grams that are relevant to the video content, demonstrating that our model generates more video-relevant captions.

Besides, we also report the results of our model on the public test set of the recent published VATEX dataset as shown in Table~\ref{vatex}. Compared with these SOTA methods, our model achieves the best performance on \textit{all metrics}. Specifically, our $D^{2}$ reaches 60.9\% in terms of CIDEr-D, which advances the strongest competitor OPENBOOK~\cite{zhang2021open} by 5.9\%, proving the effectiveness of our \emph{decoupled} pipeline.



\subsection{Ablation Analysis}
In this section, we conduct several ablation studies on the VATEX and MSRVTT datasets to determine some hyper-parameter and demonstrate the effectiveness of our proposed module. Since our method directly takes video frame as input and does end-to-end training; before doing the experiments about the main module, in section~\ref{sec:fps}, we first give the experiments about sample rates of input clips. Then, we give the main ablation experiments in section~\ref{sec:ablation_main} about our proposed \emph{Residual-Aware Temporal Block(RATB)} and \emph{Syntax-Aware Decoder(SAD)}. To fully exploit the effectiveness of the module in \emph{Residual-Aware Temporal Block(RATB)}, we did some experiments and give results in section~\ref{sec:temporal}. Finally, we give the ablation experiments results about image-text pre-training and end-to-end training in section~\ref{sec:backbone} and section~\ref{sec:e2e} respectively. 

\subsubsection{Ablation on number of input clips}
\label{sec:fps}
\begin{table}[t]
\centering
    \begin{tabular}{c cccc}           
    \toprule[1pt]
     Number of clips & \multicolumn{2}{c}{MSRVTT} & \multicolumn{2}{c}{VATEX}\\
    ~   & BLEU@4  & CIDEr-D & BLEU@4  & CIDEr-D \\ 
    \hline
    8  & 44.1    & 55.5    & 34.2 & 59.2    \\
    12  & {\bf44.5} & 56.3   & {\bf35.1}    & {\bf60.9}    \\
    16       & 44.3    & {\bf56.5}   & 34.3    & 60.9    \\
    
    \toprule[1pt]
    \end{tabular}
\caption{\textbf{\textit{Ablation on number of input clips}}.}
\label{fps}
\end{table}

In order to investigate the impact of the different number of input clips, as shown in Table ~\ref{fps}, we set the number of clips to 8, 12, 16 for our model respectively. The performance cannot increase with more clips and we use 12 for our experiments.

\subsubsection{Ablation of main design of our $\mathcal{D}^{2}$}
\label{sec:ablation_main}
\begin{table}[h!]
\centering
    \begin{tabular}{l|cc|cc|cc}
    \toprule[1pt]
    Method &
    \multicolumn{2}{c|}{Module} &
    \multicolumn{2}{c|}{MSRVTT} &
    \multicolumn{2}{c}{VATEX} \\
    \cline{2-7}
    \noalign{\smallskip}
    & RATB & SAD & B@4 & C & B@4 & C \\
    \noalign{\smallskip}\hline\noalign{\smallskip}
    Baseline &            &            &  43.3     &  52.1 & 33.9     & 55.2  \\
    Baseline & \checkmark &            & 43.9       & 54.2 & 34.2      & 57.3   \\
    Baseline &            & \checkmark & 44.1       & 53.9 & 34.4      & 57.1  \\
    Baseline & \checkmark & \checkmark & {\bf44.5}  & {\bf56.3} & {\bf35.1} & {\bf60.9}     \\
    \bottomrule[1pt]
    \end{tabular}%
\caption{\textbf{\textit{Ablation of main design of our $\mathcal{D}^{2}$}}. Spatial encoder together with vanilla decoder is selected as our baseline.}
\label{tab:my-table}
\end{table}
 We study the benefits of RATB and SAD of our method. Specifically, the spatial encoder with image-text pre-training and vanilla transformer-based decoder are combined as our baseline model. As shown in Table~\ref{tab:my-table}, our \emph{Residual-Aware Temporal Block(RATB)} boost CIDEr score by 1.9\%(2.1\%) on VATEX(MSRVTT) dataset, demonstrating the importance of building temporal dependency for video frames. Compared Row 1 and Row 3 in Table~\ref{tab:my-table}, we observe that our \emph{Syntax-Aware Decoder(SAD)} boost CIDEr-D score from 55.2(52.1) to 57.1(53.9) on VATEX(MSRVTT) dataset, proving the effectiveness of decouple the sentence generation. Combining the above two modules, the best performance(60.9 on VATEX; 56.3 on MSRVTT) is achieved on both VATEX and MSRVTT datasets.
 

\subsubsection{Ablation of Residual-Aware Temporal Block}
\label{sec:temporal}
\begin{table}[h!]
\centering
    \begin{tabular}{l|cc|cc|cc}
    \toprule[0.5pt]
    Method &
    \multicolumn{2}{c|}{Module} &
    \multicolumn{2}{c|}{MSRVTT} &
    \multicolumn{2}{c}{VATEX} \\
    \cline{2-7}
    \noalign{\smallskip}
    & RA & TA & B@4 & C & B@4 & C \\
    \noalign{\smallskip}\hline\noalign{\smallskip}
    Baseline &            &   & 42.8 & 52.1 & 33.9 & 55.2 \\
    Baseline & \checkmark & & 43.3  & 53.2 & {\bf34.3}  & 56.1     \\
    Baseline &            & \checkmark & 43.1  & 53.5 & 34.0 & 56.4    \\
    Baseline & \checkmark & \checkmark & {\bf43.9}  & {\bf54.2} & 34.2  & {\bf57.3}     \\
    \noalign{\smallskip}\noalign{\smallskip}
    \toprule[0.5pt]
    \end{tabular}
\caption{\textbf{\textit{Ablation of Residual-Aware Temporal Block}}. \textbf{\textit{RA}}, \textbf{\textit{TA}} denote \textbf{\textit{Residual-Level Attention}, \textbf{\textit{Temporal-Level Attention}}}, respectively. Spatial encoder together with vanilla decoder is selected as our baseline.}
\label{tab:RATB}
\end{table}
Compared with existing methods, our Residual-Aware Temporal Block builds temporal dependency between frames in both short-range and long-range. Specifically, \textit{Residual-Level Attention(RA)} is adopted to encode short-range(atomic action) relation. Together with \textit{Temporal-Level Attention}, atomic actions in short-range segments can be contextualized with the rest of the video. Table~\ref{tab:RATB} presents the effect of two sub-modules. The results show that \textit{RA} or TA is available as an effective sub-module and combining these two sub-modules boost the CIDEr-D score by 0.9 or 1.2(1.1 or 1.4) on VATEX(MSRVTT).

\subsubsection{Ablation of image-text pre-training}
\label{sec:backbone}
\begin{table}[t]
\centering
    \begin{tabular}{lcccc}
    \toprule[1pt]
     Init Method & \multicolumn{2}{c}{MSRVTT} & \multicolumn{2}{c}{VATEX}\\
    ~              & BLEU@4  & CIDEr-D & BLEU@4  & CIDEr-D \\ 
    \hline
    Resnet152(cls)       & 27.4 & 22.6      & 17.8    & 17.2    \\
    ViT-B/16(cls)        & 40.2 & 49.2      & 29.1    & 51.4    \\
    ViT-B/16(image-text) & {\bf43.3} & {\bf52.1} & {\bf33.9}   & {\bf55.2}    \\
    \toprule[1pt]
    \end{tabular}
\caption{\textbf{\textit{Ablation study on different type of spatial encoder and different init method.}}}
\label{init}
\end{table}
Due to the better accessibility of image-text datasets (as opposed to the video-text dataset, which is noisy and hard to collect), it is trivial for our spatial encoder to enrich visual semantics via pre-training on image-text pairs(\emph{e.g.}, MSCOCO). In this section, we study the effect of different types of spatial encoder and its init method. Results are summarized in Table~\ref{init}. We observed that there is a huge performance boost(row1 and row2) when replacing resnet152(cls) with ViT(cls), demonstrating that ViT has much more frame representation ability than the resnet. We hypothesize that it is mostly because the information loses via meanpooling operation~\cite{raghu2021vision}. Besides, when we conduct image-text pre-training on image-text dataset to our spatial encoder, the performance continues boosting, demonstrating that image-text pre-trained model benefits the video captioning task for its semantic spatial representation. 

\subsubsection{Ablation of End-to-End Training} \label{sec:e2e}
\begin{table}[h!]
    \centering
    \resizebox{0.48\textwidth}{!}{
    \begin{tabular}{lcccccc}
    \toprule[1pt]
    Method & \multicolumn{3}{c}{MSRVTT} & \multicolumn{3}{c}{VATEX} \\
           & BLEU@4 & METEOR & CIDEr & BLEU@4 & METEOR & CIDEr \\
    \midrule
    fix   &  43.6 & 29.2 & 54.8 & 34.3 &  24.4 & 58.6 \\
    e2e   &  44.5 & 30.0 & 56.3 & 35.1 &  25.1 & 60.9 \\
    \bottomrule[1pt]
    \end{tabular}
    }
    \caption{Ablation of the benefit of end-to-end training. }
    \label{tab:e2e}
\end{table}
One of the main difference between our method and previous is that we decouple the process of spatio-temporal learning: we use a 2D transformer to obtain the spatial contextual information then build temporal correlation using our designed module, giving us the benefit of end-to-end training(as opposed to 3DCNN which is hard to fine-tune). We conduct the ablation on both MSRVTT and VATEX datasets with cross-entropy loss. As it can be seen from Tab~\ref{tab:e2e}, end-to-end training boosts all metrics on MSRVTT(1.5 CIDEr score) and VATEX(2.3 CIDEr score), proving the effectiveness of our pipeline.

\subsection{Qualitative Analysis}

We show some examples in Fig.~\ref{showcase}. It can be seen, the content of captions generated by our model is richer than the baseline model, and more activity associations are involved. For instance, the example at down-left shows that the baseline model can only understand the general meaning of the video. By contrast, our model can recognize more detailed activities(\textit{e.g. } playing ping pong). The rest of the examples have similar characteristics.

\section{Conclusion}
 In this paper, we propose $\mathcal{D}^{2}$, a dual-level decoupled pure transformer pipeline for end-to-end video captioning. We address the \emph{"couple"} drawbacks on both video spatio-temporal representation learning and sentence generation. For video spatio-temporal representation, we present "first-spatial-then-temporal" paradigm. technically, we use a 2D vision transformer to generate a spatial representation for each frame and build both short and long range temporal dependency with our proposed \emph{Residual-Aware Temporal Block(RATB)}. For sentence generation, we \emph{decouple} the generation process of visual semantic and syntax-related words. Specifically, utilizing the syntactic prior provided by pre-trained language model, our proposed \emph{Syntax-Aware Decoder(SAD)} dynamically measures the contribution of visual features and syntactic prior information for the generation of each word, obtaining a more reasonable and fine-grained video captioning generation. We evaluate our methods on MSVD, MSR-VTT and VATEX, all leading to the best performance on all metrics. This sets the new state-of-the-art and our model could be the new backbone model for video captioning.

\clearpage
\bibliographystyle{ACM-Reference-Format}
\bibliography{sample-base}

\appendix

\end{document}